%% file: paper.tex
\documentclass{article}

\usepackage[preprint]{neurips_2019}

\usepackage[utf8]{inputenc} %
\usepackage[T1]{fontenc}    %
\usepackage{url}            %
\usepackage{booktabs}       %
\usepackage{amsfonts}       %
\usepackage{nicefrac}       %
\usepackage{microtype}      %

\input{include/packages.tex}
\input{include/acronyms.tex}

\input{include/commands.tex}

\title{Edge Contraction Pooling for Graph Neural Networks}

\author{%
	Frederik Diehl\\
	fortiss GmbH\\
	An-Institut Technische Universität München\\
	Munich, Germany \\
	\texttt{f.diehl@tum.de} \\
	}

\begin{document}

\maketitle

\begin{abstract}
\input{include/abstract.tex}
\end{abstract}

\input{include/introduction.tex}

\input{include/related_work.tex}

\input{include/model.tex}

\input{include/experiments.tex}

\input{include/results.tex}

\input{include/conclusion.tex}

\newpage

\bibliography{paper}
\bibliographystyle{icml2019}

\end{document}

%% file: include/packages.tex
\usepackage{todonotes}

\usepackage{glossaries}
\usepackage{booktabs}
\usepackage{dblfloatfix}    %
\usepackage{float}
\usepackage{placeins}
\usepackage{lipsum}     %
\usepackage{subcaption}
\usepackage{tikz}
\usepackage{amssymb}
\usepackage{amsmath}
\usepackage{booktabs}

\usepackage{xcolor}
\usepackage{multirow}

\usepackage{subcaption}

%% file: include/acronyms.tex
\newacronym{GCN}{GCN}{Graph Convolutional Network}
\newacronym{GNN}{GNN}{Graph Neural Network}
\newacronym{GAT}{GAT}{Graph Attention Network}
\newacronym{CNN}{CNN}{Convolutional Neural Network}
\newacronym{RNN}{RNN}{Recurrent Neural Network}
\newacronym{CVM}{CVM}{Constant Velocity Model}
\newacronym{IDM}{IDM}{Intelligent Driver Model}
\newacronym{MLP}{MLP}{Multi-Layer Perceptron}
\newacronym{HMM}{HMM}{Hidden Markov Model}

\newacronym{SAGPool}{SAGPool}{Self-Attention Graph Pooling}

\newacronym{pp}{p.p.}{percentage points}

%% file: include/commands.tex
\newcommand{\tableref}[1]{Table~\ref{#1}}
\newcommand{\figref}[1]{Fig.~\ref{#1}}
\newcommand{\equationref}[1]{Eq.~(\ref{#1})}

\newcommand{\orcidID}[1]{\href{https://orcid.org/#1}{\includegraphics[width=1em]{include/ORCID-iD_icon-64x64.png}}}

\renewcommand{\orcidID}[1]{}

\newcommand{\edgepool}{EdgePool}
\newcommand{\concatenate}{\|}

\newcommand{\softmax}{\text{softmax}}

\definecolor{deemph-gray}{gray}{0.35}

\newcommand{\datasetname}[1]{\textsc{#1}}

\newcommand{\enzymes}{\datasetname{enzymes}}
\newcommand{\proteins}{\datasetname{proteins}}
\newcommand{\redditbinary}{\datasetname{reddit-binary}}
\newcommand{\reddittwelve}{\datasetname{reddit-multi-12k}}
\newcommand{\collab}{\datasetname{collab}}

\newcommand{\cora}{\datasetname{cora}}
\newcommand{\citeseer}{\datasetname{citeseer}}
\newcommand{\pubmed}{\datasetname{pubmed}}

\newcommand{\photo}{\datasetname{photo}}
\newcommand{\computer}{\datasetname{computer}}

\newcommand{\sageconv}{SAGEConv}

\definecolor{bestcolor}{rgb}{0.8,0,0} 

\newcommand{\best}[1]{\mathbf{\textcolor{bestcolor}{#1}}}

\newcommand{\pp}{\,\gls{pp}}

%% file: include/abstract.tex
\gls{GNN} research has concentrated on improving convolutional layers, with little attention paid to developing graph pooling layers. Yet pooling layers can enable \glspl{GNN} to reason over abstracted groups of nodes instead of single nodes. To close this gap, we propose a graph pooling layer relying on the notion of edge contraction: \edgepool{} learns a localized and sparse hard pooling transform. We show that \edgepool{} outperforms alternative pooling methods, can be easily integrated into most \gls{GNN} models, and improves performance on both node and graph classification.

%% file: include/introduction.tex
\section{Introduction}

In recent years, a fast-growing field of applying deep learning to graphs has emerged. Many of these works are inspired by generalizing \glspl{CNN} to the non-euclidian and sparsely connected data that graphs represent. But while a multitude of different \glspl{GCN} have been proposed, the number of proposed pooling layers remains small. 

Yet intelligent pooling on graphs holds significant promise: It might both identify clusters (feature- or structure-based) and reduce computational requirements by reducing the number of nodes. Together, these promise to abstract from flat nodes to hierarchical sets of nodes. They are also a stepping stone towards enabling \glspl{GNN} to modify graph \textit{structures} instead of only node features.

\begin{figure}[h]
	\centering
	\includegraphics[width=1\textwidth]{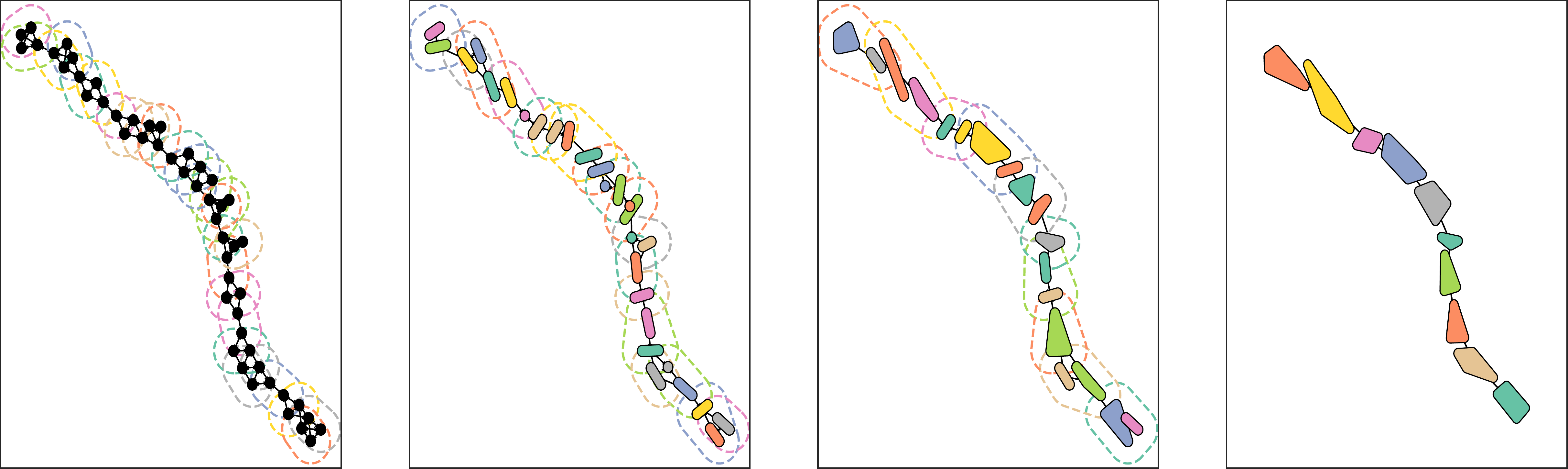}
	\caption{Edge pooling in action on a graph from \proteins{}. The original graph (left-most) is pooled three times and results in the graph depicted to the right. In each step, nodes that will be merged are surrounded by a dashed line of a random colour. In the next step, the nodes are drawn as their convex hull, filled with the same colour. Notice how the the pooled graph keeps the mostly-linear structure of the original graph. Best viewed on screen.}
	\label{fig:teaser}
\end{figure}

We propose a new pooling layer based on edge contractions (\edgepool{}, see \figref{fig:teaser}), which aims to correct weaknesses in previously proposed learned pooling layers. We do this by viewing the task not as choosing nodes but as choosing edges and pooling the connected nodes. This immediately and naturally takes the graph structure into account and ensures that we never drop nodes completely.

The main advantages of our proposed \edgepool{} layer are:

\begin{itemize}
	\item \edgepool{} performs better than other pooling methods.
	\item \edgepool{} can be integrated in existing graph classification architectures.
	\item \edgepool{} can be used for node classification and improves performance.
\end{itemize}

%% file: include/related_work.tex
\section{Related work}

Graph pooling strategies can be divided into two types: We can either use \textit{fixed} pooling methods, usually based on graph topology, or use \textit{learned} pooling methods. We concentrate on comparisons with learned pooling methods, since these appear to outperform fixed pooling methods.

\paragraph{DiffPool} \citet{yingHierarchicalGraphRepresentation2018} were the first to propose a learned pooling layer. DiffPool learns to soft-assign each node to a fixed number of clusters based on their features. DiffPool works well, but suffers from three disadvantages: (a) The number of clusters has to be chosen in advance, which might cause performance issues when used on datasets with different graph sizes. (b) Since cluster assignment is based only on node features, nodes are assigned to the same cluster based on their features, ignoring distances. (c) The cluster assignment matrix is dense, and in $\mathbb{R}^{n\times c}$, where $c$ is the number of clusters. Since $c$ is usually chosen according to the total number of nodes, the cluster assignment matrix scales quadratically with the number of nodes $n$. They also need several auxiliary objectives (link prediction, node feature $\ell_2$ regularization, cluster assignment entropy regularization) to train well. In addition to that, the density makes integration into usually sparse \glspl{GNN} difficult.

\paragraph{TopKPool} Graph U-Net, introduced by \citet{gaoGraphUNet2018a}, uses a simple top-k choice of nodes for their gPool layer, learning a node score and dropping all but the top nodes. \citet{cangeaSparseHierarchicalGraph2018a} later applied this  to graph classification. While this approach is both sparse and variable in graph size, its node choice is dependent on global state. This introduces two new issues: (a) Adding nodes to a graph can change the pooling result of the whole graph. (b) Whole areas of a graph might see no node chosen, which causes loss of information. 

\paragraph{SAGPool} \citet{leeSelfAttentionGraphPooling2019} introduced \gls{SAGPool}. A variant of TopKPool, \gls{SAGPool} no longer uses only node features to compute node scores but uses graph convolutions to take neighbouring node features into account. While their method improves TopKPool qualitatively, the disadvantages remain.

%% file: include/model.tex
\section{\edgepool{}}

For our work, we consider a graph $G = (V, E)$, where each of the $v$ nodes has $f$ features $V \in \mathbb{R}^{v \times f}$. Edges are represented as directed pairs of nodes without weights or features.

\subsection{Edge contraction}

We base our pooling operation on edge contractions. Contracting the edge $e = \{v_i, v_j\}$ introduces the new node $v_e$ and new edges such that $v_e$ is adjacent to all nodes $v_i$ or $v_j$ has been adjacent to. $v_i$, $v_j$, and all their edges are deleted from the graph. Since edge contractions are commutative, we can also define an edge set contraction. By constructing the set such that no two edges are incident to the same node, we can simply apply the naive notion of single-edge contraction multiple times.

Intuitively, we choose a single edge to contract by merging its nodes. This new node is then connected to all nodes the merged nodes had been connected to. We repeat this procedure multiple times, taking care not to include a newly-merged node into it.

\subsection{Choosing edges}

Given the preconditions mentioned above, we naively choose edges by computing a score for each edge, then iteratively contracting the highest-scoring edge which does not have a newly-merged node incident.

In our procedure, we compute raw scores for each node as a simple linear combination of the concatenated node features. For an edge from node $i$ to node $j$, we compute the raw score $r$ as 

\begin{equation}
	r(e_{ij}) = W \cdot  (n_i \concatenate n_j) + b,
	\label{eq:raw_edge_scores}
\end{equation}

where $n_i$ and $n_j$ are the node features and $W$ and $b$ are learned parameters.

To compute the final node score $s_{ij}$ for an edge, we employ a local softmax normalization over all edges of a node\footnote{We experimented with a simple $\tanh$ gating function, but found softmax normalization to perform better.}. We modify the final score such that the mean of the score range lies at $1$. Later on, this enables us to include the score in the unpooling procedure without issues due to numerical stability. We also found this to lead to better performance in the graph classification task, which we believe is because of better gradient flow. The final score then becomes:

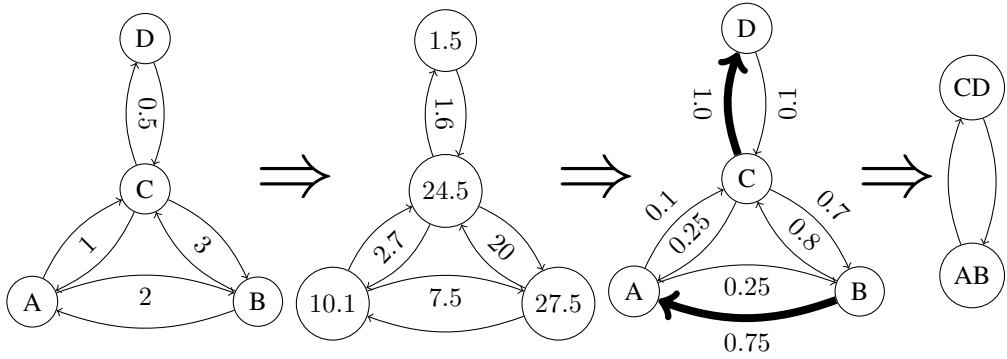
\begin{figure}
	\centering
	\input{imgs/explain_softmax.tikz}
	\caption{Edge score computation. To reduce visual clutter, we assume both directions of a node to have the same  raw score. We also greatly round numbers. (a) The raw scores for each edge. (b) This shows the computed $\exp{(r)}$ for each edge and the sum over all incoming raw scores for each node. (c) Final scores for each edge. The edges chosen to be contracted are marked bold. (d) Resulting, pooled, graph. As can be seen, edge BC was not contracted because node C is incident to the previously contracted CD edge.}
	\label{fig:explain_softmax}
\end{figure}

\begin{equation}
	s_{ij} = 0.5 + \softmax_{r_{*j}} (r_{ij}).
\end{equation}

Given the edge scores, we now iteratively contract edges according to the scores, ignoring those which have a newly-merged node incident. An illustration of the process is depicted in \figref{fig:explain_softmax}.

Note that this will always pool roughly 50\% of the total nodes. Contrary to DiffPool and TopKPool, this ratio cannot be changed.

\subsection{Computing new node features}

There are many strategies for combining the features of pairs of nodes. In particular, we are not restricted  to symmetric functions since the edges chosen have a specific direction. Nonetheless, we found that taking the sum of the node features works well.

To enable the gradient to flow into the scores, we use gating and multiply the combined node features by the edge score:

\begin{equation}
	\hat{n}_{ij} = s_{ij} \left( n_i + n_j\right)
\end{equation}

\subsection{Computational performance}
Given our procedure above, we immediately see that \edgepool{} can operate on sparse representations. When doing so, both runtime and memory scales linearly in the number of edges. This particularly avoids the scaling issues of DiffPool's cluster assignment matrix.

Additionally, \edgepool{} is locally independent: As long as the node scores of two nodes $n_i$ and $n_j$ and of their neighbours do not change (by changing nodes within the receptive fields), the choice of edge $e_{ij}$ will not change. Accordingly, \edgepool{} does not have to be computed for the whole graph at once. If the graph changes, only the pooling local to the changed areas needs to be updated.

\subsection{Integrating edge features}

\edgepool{} can be updated to take edge features $f_{ij}$ of edge $e_{ij}$ into account. To do so, we have to include them in the raw score computation (\equationref{eq:raw_edge_scores}). The simplest approach is to concatenate them:

\begin{equation}
r(e_{ij}) = W \cdot  (n_i \concatenate n_j \concatenate f_{ij}) + b.
\end{equation}

Additionally, we will likely have to change the procedure to compute new node features; we propose using a weighted linear combination of both nodes' features, the features of the chosen edge, and the features of the reverse edge if it exists.

Lastly, we need a procedure to combine the edge features of edges that ended at both merged nodes and will therefore be merged. We believe a simple sum should work well here, too. However, we have not conducted experiments on this.

\subsection{Unpooling \edgepool{}}

To use pooling in the context of node classification, an unpooling operation is necessary. To do so, each \edgepool{} layer also emits the mapping of each of the previous graph's nodes to the newly-pooled graph's nodes. When unpooling, we then create an inverse mapping of pooled nodes to unpooled nodes. Since we assign each node to exactly one merged node, this mapping can be chained through many pooling layers. Additionally, we divide the unpooled node features by the corresponding edge score:

\begin{equation}
	\hat{n}_{i} = \hat{n}_{j} = n_{ij} / s_{ij}.
\end{equation}

%% file: imgs/explain_softmax.tikz
 
\begin{tikzpicture}
	\node at (0, 0) {
		\begin{tikzpicture}[chosen/.style={line width=3pt}]

		\node[draw,circle] (A) at (-1.5,0) {A};
		\node[draw,circle] (B) at (1.5,0) {B};
		\node[draw,circle] (C) at (0,1.5) {C};
		\node[draw,circle] (D) at (0,3.5) {D};		
		
		\draw [->] (A) to [bend left=20] node[sloped,pos=0.5,below=0.05cm] {$2$} (B);
		\draw [->] (B) to [bend left=20] node[sloped,pos=0.5,below=0.05cm] {} (A);
		\draw [->] (C) to [bend left=20] node[sloped,pos=0.5,above=0.05cm] {$1$} (A);
		\draw [->] (A) to [bend left=20] node[sloped,pos=0.5,above=0.05cm] {} (C);
		\draw [->] (B) to [bend left=20] node[sloped,pos=0.5,above=0.05cm] {$3$} (C);
		\draw [->] (C) to [bend left=20] node[sloped,pos=0.5,above=0.05cm] {} (B);
		\draw [->] (C) to [bend left=20] node[sloped,pos=0.5,above=0.05cm] {$0.5$} (D);
		\draw [->] (D) to [bend left=20] node[sloped,pos=0.5,below=0.05cm] {} (C);

		\end{tikzpicture}
		};
	\node[scale=3] at (2, 0) {$\Rightarrow$};
	
	\node at (4, 0) {
		\begin{tikzpicture}[chosen/.style={line width=3pt}]
		
		\node[draw,circle] (A) at (-1.5,0) {$10.1$};
		\node[draw,circle] (B) at (1.5,0) {$27.5$};
		\node[draw,circle] (C) at (0,1.5) {$24.5$};
		\node[draw,circle] (D) at (0,3.5) {$1.5$};

		\draw [->] (A) to [bend left=20] node[sloped,pos=0.5,below=0.05cm] {$7.5$} (B);
		\draw [->] (B) to [bend left=20] node[sloped,pos=0.5,below=0.05cm] {} (A);
		\draw [->] (C) to [bend left=20] node[sloped,pos=0.5,above=0.05cm] {$2.7$} (A);
		\draw [->] (A) to [bend left=20] node[sloped,pos=0.5,above=0.05cm] {} (C);
		\draw [->] (B) to [bend left=20] node[sloped,pos=0.5,above=0.05cm] {$20$} (C);
		\draw [->] (C) to [bend left=20] node[sloped,pos=0.5,above=0.05cm] {} (B);
		\draw [->] (C) to [bend left=20] node[sloped,pos=0.5,below=0.05cm] {} (D);
		\draw [->] (D) to [bend left=20] node[sloped,pos=0.5,below=0.05cm] {$1.6$} (C);
		\end{tikzpicture}
	};
	
	\node[scale=3] at (6, 0) {$\Rightarrow$};
		
	\node at (8, 0) {
		\begin{tikzpicture}[chosen/.style={line width=3pt}]
		
		\node[draw,circle] (A) at (-1.5,0) {A};
		\node[draw,circle] (B) at (1.5,0) {B};
		\node[draw,circle] (C) at (0,1.5) {C};
		\node[draw,circle] (D) at (0,3.5) {D};		
		
		\draw [->] (A) to [bend left=20] node[sloped,pos=0.5,below=0.05cm] {$0.25$} (B);
		\draw [->,chosen] (B) to [bend left=20] node[sloped,pos=0.5,below=0.05cm] {$0.75$} (A);
		\draw [->] (C) to [bend left=20] node[sloped,pos=0.5,above=0.05cm] {$0.25$} (A);
		\draw [->] (A) to [bend left=20] node[sloped,pos=0.5,above=0.05cm] {$0.1$} (C);
		\draw [->] (B) to [bend left=20] node[sloped,pos=0.5,above=0.05cm] {$0.8$} (C);
		\draw [->] (C) to [bend left=20] node[sloped,pos=0.5,above=0.05cm] {$0.7$} (B);
		\draw [->,chosen] (C) to [bend left=20] node[sloped,pos=0.5,below=0.05cm] {$1.0$} (D);
		\draw [->] (D) to [bend left=20] node[sloped,pos=0.5,below=0.05cm] {$0.1$} (C);

		\end{tikzpicture}
		};
		
	\node[scale=3] at (10, 0) {$\Rightarrow$};
	
	\node at (11, 0) {
		\begin{tikzpicture}[chosen/.style={line width=3pt}]
		
		\node[draw,circle] (AB) at (0,0) {AB};
		\node[draw,circle] (CD) at (0,2.5) {CD};
		
		\draw [->] (AB) to [bend left=20] node[sloped,pos=0.5,below=0.05cm] {} (CD);
		\draw [->] (CD) to [bend left=20] node[sloped,pos=0.5,below=0.05cm] {} (AB);		
		\end{tikzpicture}
	};
\end{tikzpicture}

%% file: include/experiments.tex
\section{Experiments}
\label{sec:experiments}

We design our experiments to answer three questions:

\begin{description}
	\item[Q1:] Does \edgepool{} outperform alternative pooling methods?
	\item[Q2:] Can \edgepool{} be used as a plug-and-play addition for any \gls{GNN}?
	\item[Q3:] Can \edgepool{} be used for node classification?
\end{description}

\subsection{General Setup}

We evaluate our models on multiple graph and node classification datasets, and share most of the training procedures between all models.

\subsubsection{Datasets}

While there are many graph classification datasets available, most of these are small (in both nodes per graph and total graphs). As an example, the popular \enzymes{} dataset contains only 600 graphs, making 10-fold crossvalidation (at a test set size of 60) very difficult.

We conduct 10-fold cross-validation for all datasets and report mean and standard deviation. We choose all folds at random, eschewing the default planetoid split.

\paragraph{Graph classification datasets}

For graph classification, we evaluate on four datasets from the collection by \citet{KKMMN2016}. At 1113 graphs, \proteins{} \citep{borgwardtProteinFunctionPrediction2005} is the smallest, but has been used extensively as a benchmark dataset. The task is to predict whether a given protein is an enzyme. The two reddit-based datasets \citep{yanardagDeepGraphKernels2015} depict user responses in an online discussion. The task is to predict the subreddit, out of two (\redditbinary{}) or eleven (\reddittwelve{}) choices. Lastly, each \collab{} (ibid) graph models scientific collaborations of one researcher. The task is to classify which of three fields the researcher belongs to. Neither \collab{} nor the two reddit-based datasets have any features.

\paragraph{Node classification datasets}
We also evaluate \edgepool{} on five semi-supervised node classification datasets. \cora{} \citep{namataQuerydrivenActiveSurveying2012}, \citeseer{}, and \pubmed{} \citep{senCollectiveClassificationNetwork2008} model citation networks. In these, nodes are documents and edges model citations. The goal is to classify the subfield of each of the documents. The \photo{} and \computer{} datasets \citep{shchurPitfallsGraphNeural2018} are part of the Amazon co-purchasing graph. Nodes are products and edges model co-purchases between products. The goal is to predict the product category.

Each of these datasets is a semi-supervised node classification task from bag-of-word features. We use 20 nodes per class as training data and 30 nodes per class as test data. Every other node is unlabelled.

\subsection{Training}

While we use different models, several setup parameters have been chosen identically between all models. Each is trained for a total of 200 epochs using the Adam optimizer \citep{Kingma2014} with a learning rate of $10^{-3}$, which is halved every 50 epochs. 128 graphs are batched together at each step by treating them as a single unconnected graph. We use 128 channels except for \proteins{} and the node classification datasets, where we used 64. This setup follows \citet{yingHierarchicalGraphRepresentation2018}.

All models use both dropout and batch normalization \citep{Ioffe2015}. We found batch normalization to suffer greatly when evaluated using population statistics and instead use mini-batch statistics even during testing.

We also found using edge score dropout significantly increased \edgepool{}'s performance, and set every edge score to $0$ with a chance of $0.2$.

\subsection{Experimental design}

To answer the questions we have posed, we design three different experiments.

\subsubsection{Q1: Does \edgepool{} outperform alternative pooling approaches?}
To evaluate this, we use the same architecture as used by \citet{yingHierarchicalGraphRepresentation2018} for DiffPool: The model has three \sageconv{} blocks \citep{hamiltonInductiveRepresentationLearning2017} whose outputs are globally mean-pooled and concatenated. Final classification occurs after two fully-connected layers. The base model does not pool nodes, every other model pools after every block. Note that DiffPool uses a siamese architecture, using separate \sageconv{} blocks to compute cluster assignments. We restrict DiffPool to a maximum of 750 nodes per graph and set TopKPool's pool ratio to 0.5 to remain comparable to EdgePool.

Additionally, we only use the cross-entropy loss to train the model. To ensure a fair comparison, we also do this for DiffPool, which originally used three additional auxiliary losses and tasks to stabilize training and precomputed additional features.

\subsubsection{Q2: Can \edgepool{} be integrated in existing architectures?}
To evaluate whether \edgepool{} can be integrated into pre-existing architectures, we follow the model configuration from \textit{pytorch-geometric}'s benchmarks \citep{feyFastGraphRepresentation2019}. Speficially, we use a total of seven convolutional layers, followed by a global pooling layer and two fully-connected layers. If pooling is used, it is added after every second convolutional layer (i.e. there are three pooling layers).

The convolutional layers we evaluate this on are GCN \citep{kipf_semi-supervised_2016}, GIN and GIN0 \citep{xu2018how}, and GraphSAGE \citep{hamiltonInductiveRepresentationLearning2017} both with and without accumulating intermediate results (SAGE nacc). Additionally, we construct a model using node-independent \glspl{MLP}, in which only pooling might lead to communication between nodes.

\subsubsection{Q3: Can \edgepool{} be used for node classification?}
On node classification tasks, we evaluate a simple architecture, varying the convolutional layers. We evaluate GCN, GIN and GIN0, and GAT \citep{velickovicGraphAttentionNetworks2017}. Again, we also evaluate a \glspl{MLP} layer. As with Q2, we use seven convolutional layers. We pool after the second and fourth and unpool after the fifth and seventh, with shortcuts between the poolings. The concatenated features are then used by a two-layer MLP to predict each node's class.

%% file: include/results.tex
\section{Results and discussion}

We implemented the models using PyTorch \citep{paszke2017automatic} and in particular the pytorch-geometric library \citep{feyFastGraphRepresentation2019}. Experiments were conducted on several Geforce 1080Ti GPUs in parallel, leveraging Singularity containers \citep{Kurtzer2017} for reproducibility.

\subsection{\edgepool{} vs. alternative pooling approaches}

\tableref{tab:results_pooling} shows mean accuracy and standard deviation for graph classification tasks. As can be seen, \edgepool{} consistently improves performance over the non-pooling models and TopKPool. Discounting \proteins{} due to close performance, it outperforms all other pooling approaches on two tasks, and is only outperformed by DiffPool on one task.

This answer \textbf{Q1}: \edgepool{} consistently outperforms all pooling methods but DiffPool. While DiffPool might perform better on some graphs, \edgepool{} scales far better and can be used on large graph sizes.

\begin{table}
	\centering
	\input{tables/results_pooling.tex}
	\label{tab:results_pooling}
\end{table}

\subsection{\edgepool{} in existing architectures}

\tableref{tab:results_benchmarks} shows comparative results for different benchmark models with and without \edgepool{}. On a large majority of \gls{GNN}/dataset combinations, \edgepool{} increases performance, by an average of almost $2$\pp{}. GIN and GIN0 profit the least (mean improvement of $0.3\pp{}$), while GraphSAGE profits the most ($5.5$\pp{}).

Interestingly, we can see that \edgepool{} allows even the \gls{MLP} model to perform fairly well. This model can only rely on pooling to gain information on the neighbourhood. Nonetheless, it performs competitively on \proteins{} and \collab{}.

Unfortunately, the performance increases of \edgepool{} are not consistent over datasets and models. This makes it impossible to make a specific recommendation on situations in which one should or should not include \edgepool{} in the model.

However, we can still answer \textbf{Q2}: It is easily possible to integrate \edgepool{} in existing architectures. Doing so will lead to an estimated improvement of about $2$\pp{}, but might for some combinations of model and dataset decrease performance.

\begin{table}
	\centering
	\input{tables/results_benchmarks.tex}
	\label{tab:results_benchmarks}
\end{table}

\subsection{\edgepool{} for node classification}

As \tableref{tab:results_benchmarks_node} shows, \glspl{GNN} using \edgepool{} can be integrated in node classification architectures and improves performance for 21 of 25 dataset/model combinations.

In particular, note the increase in performance for the \gls{MLP}. In several of these tasks, an \gls{MLP} augmented with \edgepool{} shows competitive performance to \gls{GNN} algorithms. For \gls{GNN} algorithms, \edgepool{} improves performance by an average of $3.5$\pp{}, performing worst on \pubmed{} (no improvement on average) and for \glspl{GCN} (decrease by $0.1$\pp{}). It performs best for GIN and GIN0, at $5.8$\pp{} and $6.6$\pp{} improvements respectively.

This answers \textbf{Q3}: \edgepool{} will, in most cases, improve performance for node classification. The expected improvement is an average of $3.5$\pp{}.

\begin{table}
	\centering
	\input{tables/results_node_benchmarks.tex}
	\label{tab:results_benchmarks_node}
\end{table}

\subsection{Visual inspection}

\begin{figure}
	\centering
	\begin{subfigure}[b]{1.0\textwidth}
	\centering
	\includegraphics[width=1\textwidth]{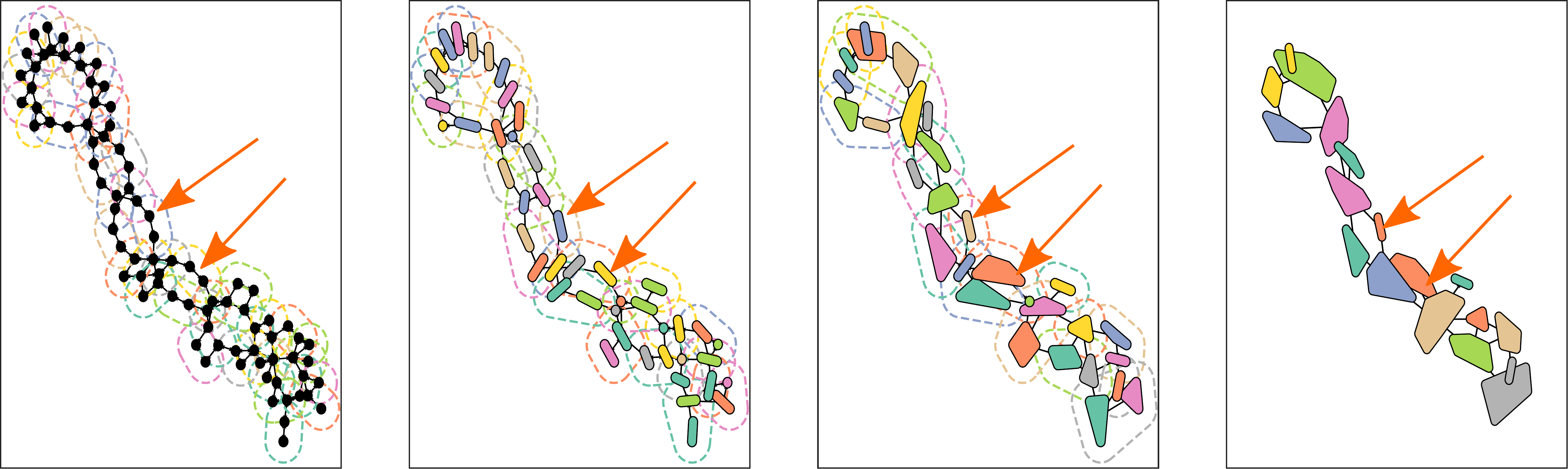}
	\caption{\edgepool{} example. The graph is simplified in each step. In particular, note the two node groups marked by the orange arrows. Locally, these form two unconnected paths which stay unconnected throughout the pooling process.}
	\label{fig:pooling_example}
	\end{subfigure}

	\begin{subfigure}[b]{1.0\textwidth}
	\centering
	\includegraphics[width=1\textwidth]{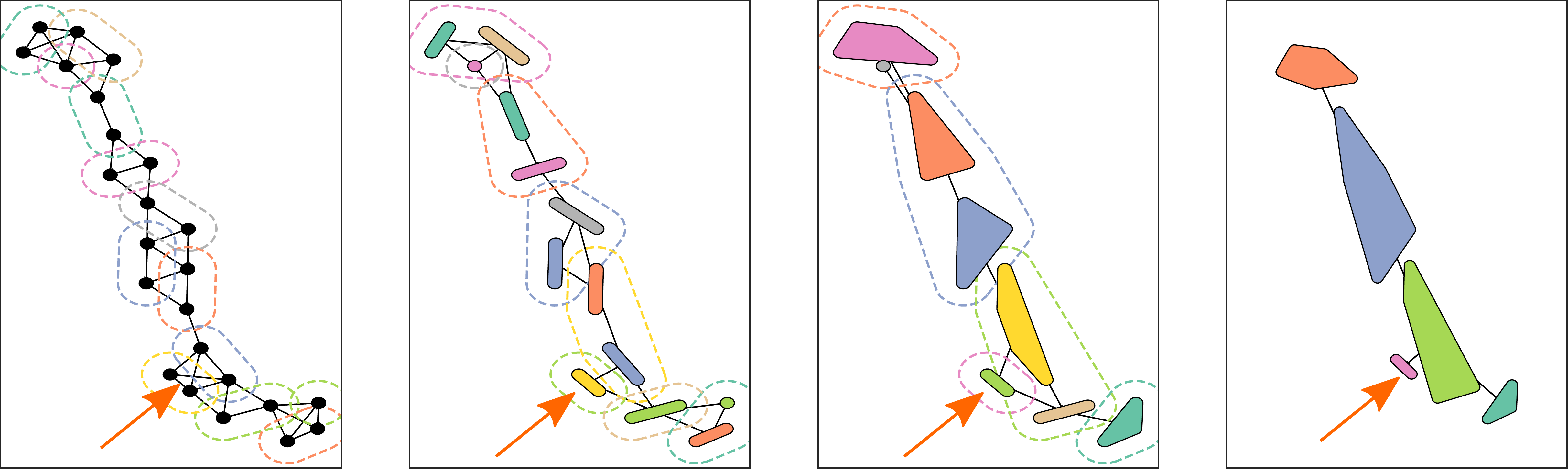}
	\caption{In this case, \edgepool{} fails to pool human-visible structures together. The two nodes marked by the orange arrow are closely connected to their neighbours. In the final pooling configuration, however, they have not been joined with them and remain separate.}
	\label{fig:failure_case}
	\end{subfigure}
	\caption{Edge pooling on graphs from \proteins{}. Visualization is identical to \figref{fig:teaser}. Best viewed on screen and zoomed in.}
	\label{fig:plotted_graphs}
\end{figure}

Both \figref{fig:teaser} and \figref{fig:plotted_graphs} show examples of the pooling resulting from using \edgepool{}. In particular, they show that \edgepool{} keeps the linearity of the original protein even after pooling. \figref{fig:pooling_example} shows how unconnected paths (orange arrows) are not pooled, keeping the original graph structure visible even after pooling. However, as \figref{fig:failure_case} shows, there are situations in which \edgepool{} causes node poolings which are counter-intuitive to humans.

%% file: tables/results_pooling.tex
\caption{Comparing pooling strategies: Accuracy (and standard deviation) on benchmark datasets in percent. Best results are marked bold. [*] \citet{yingHierarchicalGraphRepresentation2018} use several additional techniques and auxiliary losses to stabilize training, and also include additional computed features. We report results without these.} 

\begin{tabular}{lrrrr}
	\toprule
	                   & \textbf{\datasetname{proteins}} & \textbf{\datasetname{rdt-b}} & \textbf{\datasetname{rdt-12k}} & \textbf{\datasetname{collab}}\\ \midrule
	Base Model         & $71.4\pm 3.2$ 			& $69.9\pm 3.7$ 		& $35.1\pm 1.6$ 		& $65.4\pm 1.5$ \\
	DiffPool [*]       & $72.3\pm 5.8$ 	        & $82.9\pm 3.4$ 		& $34.8\pm 1.9$ 		& $\best{70.1\pm 1.5}$\\
	TopKPool           & $70.6\pm 4.8$ 			& $68.9\pm 3.2$ 		& $28.7\pm 1.8$ 		& $64.6\pm 2.1$\\
	SAGPool            & $71.8\pm 6.0$ 			& $84.7\pm 4.4$ 		& $41.9\pm 3.3$ 		& $63.9\pm 2.5$\\
	\edgepool{}        & $\best{72.5\pm 3.2}$ 			& $\best{87.3\pm 4.1}$ 	        & $\best{45.6\pm 1.8}$ 	& $67.1 \pm 2.7$ \\\bottomrule
\end{tabular}

%% file: tables/results_benchmarks.tex
\caption{Integrating \edgepool{} into existing architectures: Accuracy (in percent) of benchmark models with and without \edgepool{}. SAGE is short for GraphSAGE; nacc means without accumulating results.}

\begin{tabular}{lrrrrrr}
	\toprule
	\textbf{\datasetname{proteins}} & GCN              & GIN                   & GIN0                      & SAGE                  & SAGE nacc              & MLP \\ \midrule
	No Pooling                 & $71.4 \pm 5.0$        & $70.4 \pm 2.7$        & $70.9 \pm 73.7$           & $71.7 \pm 3.6$        & $\best{73.0 \pm 4.8}$ & $71.8 \pm 4.2$\\
	\edgepool{}                & $\best{73.1 \pm 4.6}$ & $\best{72.9 \pm 3.6}$ & $\best{71.7 \pm 3.6}$     & $\best{73.5 \pm 3.5}$ & $69.9 \pm 4.9$        & $\best{73.1 \pm 4.6}$\\
	
	\toprule
	\textbf{\datasetname{RDT-B}} &&&&&&\\ \midrule
	No Pooling                 & $87.1 \pm 2.8$        & $91.9 \pm 1.7$        & $92.3 \pm 1.5$            & $62.5 \pm 4.8$        & $50.3 \pm 8.4$        & $51.0 \pm 4.3$\\
	\edgepool{}                & $\best{87.8 \pm 3.1}$ & $\best{92.1 \pm 2.2}$ & $\best{93.0 \pm 1.7}$     & $\best{68.0 \pm 5.3}$ & $\best{64.5 \pm 4.6}$ & $\best{69.9 \pm 2.8}$\\
	
	\toprule
	\textbf{\datasetname{RDT-12K}} & &&&&& \\ \midrule
	No Pooling                 & $\best{47.6 \pm 0.6}$ & $\best{49.5 \pm 1.1}$ & $\best{50.0 \pm 1.3}$     & $22.9 \pm 2.3$        & $24.4 \pm 1.4$        & $21.9 \pm 1.5$\\
	\edgepool{}                & $47.4 \pm 2.1$        & $49.3 \pm 1.1$        & $49.6 \pm 1.2$            & $\best{36.9 \pm 2.1}$ & $\best{37.8 \pm 2.0}$ & $\best{34.6 \pm 1.3}$\\
	
	\toprule
	\textbf{\datasetname{COLLAB}} & &&&&&\\ \midrule
	No Pooling                 & $67.0 \pm 2.2$   & $\best{74.2 \pm 1.8}$ & $\best{74.1 \pm 1.6}$ & $63.6 \pm 2.4$ & $64.1 \pm 2.1$ & $52.0 \pm 2.5$\\
	\edgepool{}                & $\best{71.5 \pm 2.0}$   & $73.0 \pm 2.1$ & $72.2 \pm 1.6$      & $\best{64.3 \pm 1.9}$ & $\best{64.1 \pm 2.3}$ & $\best{67.8 \pm 3.2}$\\

\end{tabular}

%% file: tables/results_node_benchmarks.tex
\caption{Using \edgepool{} for node classification. Accuracy (in percent) of benchmark models with and without \edgepool{}.}

\begin{tabular}{lrrrrr}
	\toprule
	\textbf{\cora{}} & GCN & GIN         & GIN0 & GAT & MLP \\ \midrule
	No Pooling & $ 71.8 \pm 3.4$   & $ 52.1 \pm 4.7$ & $ 55.9 \pm 4.4$ & $68.0 \pm 4.5$ & $35.6 \pm 2.6$ \\
	\edgepool{} & $\best{72.8 \pm 1.9}$   & $ \best{63.0 \pm 5.4}$ & $\best{61.3 \pm 3.9}$      & $\best{70.3 \pm 3.3}$ & $\best{58.3 \pm 3.6}$\\
	
	\toprule
	\textbf{\citeseer{}} & &&&& \\ \midrule
	No Pooling & $ 62.9 \pm 2.9$   & $ 40.9 \pm 4.6$ & $ 41.4 \pm 3.8$ & $58.9 \pm 2.8$ & $35.5 \pm 3.2$ \\
	\edgepool{} & $\best{65.3 \pm 2.7}$   & $ \best{50.6 \pm 3.9}$ & $\best{49.9 \pm 5.7}$      & $\best{61.0 \pm 3.4}$ & $\best{50.0 \pm 3.7}$\\
	
	\toprule
	\textbf{\pubmed{}} & &&&& \\ \midrule
	No Pooling & $\best{74.2 \pm 1.7}$   & $60.8 \pm 6.8$ & $61.0 \pm 4.4$ & $\best{73.0 \pm 2.0}$ & $62.4 \pm 4.1$\\
	\edgepool{} & $74.1 \pm 2.1$   & $\best{61.0 \pm 6.4}$ & $\best{61.9 \pm 4.9}$      & $72.0 \pm 4.7$ & $\best{64.8 \pm 3.2}$\\
	
	\toprule
	\textbf{\photo{}} & &&&& \\ \midrule
	No Pooling & $\best{88.4 \pm 2.2}$   & $69.9 \pm 3.2$ & $71.9 \pm 4.0$ & $78.5 \pm 4.5$ & $59.6 \pm 4.9$\\
	\edgepool{} & $86.5 \pm 0.8$   & $\best{77.1 \pm 1.8}$ & $\best{78.1 \pm 1.5}$      & $\best{81.0 \pm 4.2}$ & $\best{81.4 \pm 2.3}$\\
	
        \toprule
	\textbf{\computer{}} & &&&& \\ \midrule
	No Pooling & $\best{80.0 \pm 2.6}$   & $53.1 \pm 5.5$ & $52.4 \pm 3.6$ & $60.6 \pm 12.4$ & $43.0 \pm 6.7$\\
	\edgepool{} & $77.9 \pm 2.2$   & $\best{58.1 \pm 4.8}$ & $\best{60.4 \pm 4.3}$      & $\best{62.5 \pm 13.0}$ & $\best{69.4 \pm 2.3}$\\

\end{tabular}

%% file: include/conclusion.tex
\section{Conclusion}

We have proposed \edgepool{}, a hard pooling method for \acrlongpl{GNN}, based on edge contraction.

This pooling is both localized (and therefore independent of non-local graph changes) and sparse (and therefore computationally efficient even on large graphs). 

Except for a single pooling procedure on a single dataset, \edgepool{} outperforms all previously proposed pooling approaches. We also show that \edgepool{} can be integrated into a large number of \gls{GNN} architectures and usually improves performance on both node and graph classification tasks without any adaptions to training or architecture.

Besides the obvious use of \edgepool{} in improving existing \gls{GNN} architectures, we hope it will serve as a stepping stone towards methods that learn how to modify graph structures. We believe this will lead towards methods that no longer operate on nodes but on abstracted groups of nodes.